\documentclass[11pt, a4paper]{article}
\usepackage{coling2020}
\usepackage{times}
\usepackage{url}
\usepackage{latexsym}
\usepackage[colorlinks]{hyperref}
\usepackage{graphicx}
\usepackage{caption}
\usepackage{subcaption}
\usepackage{multirow}
\usepackage{makecell}
\usepackage{booktabs}

\colingfinalcopy 

\title{Hierarchical Multitask Learning Approach for BERT}

\author{Çağla Aksoy \qquad Alper Ahmetoğlu \qquad Tunga Güngör \\
 Department of Computer Engineering, Boğaziçi University \\
 Bebek Istanbul 34342 Turkey \\
 \tt{\{cagla.aksoy,alper.ahmetoglu,gungort\}@boun.edu.tr}}

\date{17.10.2020}

\begin{document}
\maketitle

\begin{abstract}
Recent works show that learning contextualized embeddings for words is beneficial for downstream tasks. BERT is one successful example of this approach. It learns embeddings by solving two tasks, which are masked language model (masked LM) and the next sentence prediction (NSP). The pre-training of BERT can also be framed as a multitask learning problem. In this work, we adopt hierarchical multitask learning approaches for BERT pre-training. Pre-training tasks are solved at different layers instead of the last layer, and information from the NSP task is transferred to the masked LM task. Also, we propose a new pre-training task bigram shift to encode word order information. We choose two downstream tasks, one of which requires sentence-level embeddings (textual entailment), and the other requires contextualized embeddings of words (question answering). Due to computational restrictions, we use the downstream task data instead of a large dataset for the pre-training to see the performance of proposed models when given a restricted dataset. We test their performance on several probing tasks to analyze learned embeddings. Our results show that imposing a task hierarchy in pre-training improves the performance of embeddings.

\end{abstract}

\section{Introduction}
\label{intro}

There have been many studies in natural language processing (NLP) to find suitable word representations (embeddings) that carry information of a language. Even if finding these word representations can be computationally demanding, this can be advantageous since it is computed only once. These learned representations can be used for various downstream tasks.

Word2Vec \cite{mikolov2013efficient} finds word embeddings by predicting a word given its neighborhood (CBOW) or predicting its neighborhood given the word (Skip-gram). Words that are used together have similar word embeddings due to the training strategy. However, these embeddings do not contain word order information and contextual information. ELMo \cite{peters2018deep} uses bidirectional LSTM (BiLSTM) \cite{hochreiter1997long} to predict a word given its context. Since BiLSTM is used for creating embeddings, both left-to-right and right-to-left contexts are implicitly encoded. Transformer \cite{vaswani2017attention} is shown to be more appropriate for training in large datasets due to its self-attention mechanism. OpenAI GPT \cite{radford2018improving} has the same objective as ELMo in the forward direction, except it uses transformer architecture. BERT \cite{devlin2018bert} also uses transformer architecture with bidirectional pre-training tasks. Training objectives affect the information encoded in embeddings. Each objective and architecture presumes a different inductive bias.

In this work, we focused on BERT as it uses multiple training objectives. These objectives can create an inhibitory effect or a regulatory effect on each other. For this reason, we applied a hierarchical multitask learning approach to BERT by modifying its original structure. Our motivation is to create embeddings that encode the information from each task in a balanced way. Our contributions are as follows:

\begin{itemize}

\item Instead of training masked language model (masked LM) and next sentence prediction (NSP) classifiers with the last layer embeddings, we trained masked LM classifiers with embeddings from lower layers of the transformer (Lower Mask).  We do the same experiment for the NSP classifier as well (Lower NSP). By evaluating the performance of the embeddings on downstream tasks, we provide insights about the hierarchy between pre-training tasks.
\item We incorporate the input or the output of the NSP classifier to the input of the masked LM classifier in order to enrich the sentence-level embedding.
\item We propose a new pre-training objective, bigram shift,  in addition to masked LM and NSP tasks to enforce embeddings also to learn word order information.

\end{itemize}

Our experimental results show that Lower NSP has a competitive performance when compared with the original BERT structure. We also evaluate the learned embeddings on probing tasks to provide useful insights into training strategies. Results on probing task experiments show that using bigram shift task for pre-training is useful for specific tasks. The remaining part of this paper is organized as follows. In Section \ref{related}, we mention related works. In Section \ref{methods}, we explain our methods in detail. In Section \ref{experiments}, we report our experiment results. Lastly, we give a conclusion in Section \ref{conclusion}.

\section{Related Work}
\label{related}

Multitask learning (MTL) \cite{caruana1997multitask} is an approach to construct one model for related tasks to achieve a better generalization performance. BERT pre-training can be thought of as a multitask model by this definition. Besides this parallel task structure as in BERT, hierarchical MTL approach models tasks in a successive fashion. With this approach, complex tasks are predicted at deeper layers so that they can take advantage of representations of lower-level tasks \cite{sogaard2016deep,hashimoto2016joint,sanh2019hierarchical}. Examples of this approach can also be seen in computer vision \cite{szegedy2016rethinking,sabour2017dynamic}. The other important issue is choosing related tasks. In case of training a model for the tasks which do not support each other, the accuracy for each task can be lower than their single task structures \cite{bingel2017identifying}.

BERT pre-training is done by predicting randomly masked words and predicting whether two sentences are consecutive or not. These two objectives are optimized simultaneously. This model is trained on a large unlabeled corpus; it can be easily fine-tuned on various downstream tasks with high performance. There are different studies to replicate and improve BERT model architecture. RoBERTa \cite{liu2019roberta} makes slight adjustments for pre-training tasks. One relevant change to our work is removing the NSP task. This approach also hints to the issue of choosing related tasks \cite{bingel2017identifying}. ALBERT \cite{lan2019albert} proposes a way to reduce the parameter size of the BERT structure. They also proposed a new pre-training task, sentence order prediction,  instead of NSP. StructBERT \cite{wang2019structbert} also proposes two auxiliary tasks that are word structure and sentence structure instead of NSP to create embeddings that are faithful to both word and sentence orders. Word structural task is similar to our proposed bigram shift task. However, we predict whether words are in the correct position or not instead of predicting the correct word in a position.

In general, BERT is fine-tuned on downstream tasks by using its last layer activations. There are several works that use lower (intermediate) layers of BERT as feature extractors and aggregate them with additional layers \cite{yang2019deepening,zhu2018sdnet}. These works only focus on using lower layer activations in the fine-tuning step without changing the pre-training procedure. Our work differs in the sense that we modify pre-training of BERT with hierarchical multitask learning to get lower layers that are specialized on one task.

\section{Methods}
\label{methods}

In the original BERT pre-training, there are two objectives: masked LM and NSP. These are optimized simultaneously to learn a language model. From a multitask learning point of view, minimizing the loss of two objectives at the same time should enhance the performance if these two tasks are related to each other \cite{bingel2017identifying}. To our knowledge, there is no prior work on the relation between these two tasks. As the relation of tasks is an essential factor, another important factor is the task complexity. For example, if one of these tasks is more straightforward than the other, the model may overfit to easier task while the training for the other task is not complete. This is an unwanted consequence since a part of the model memorizes the task instead of generalizing it. This could have been prevented by early stopping; however, it is hard to decide when to stop even for one task \cite{prechelt1998early}.

\begin{figure}[htbp]
\centering
 \begin{subfigure}[b]{0.48\textwidth}
     \centering
     \includegraphics[width=\textwidth]{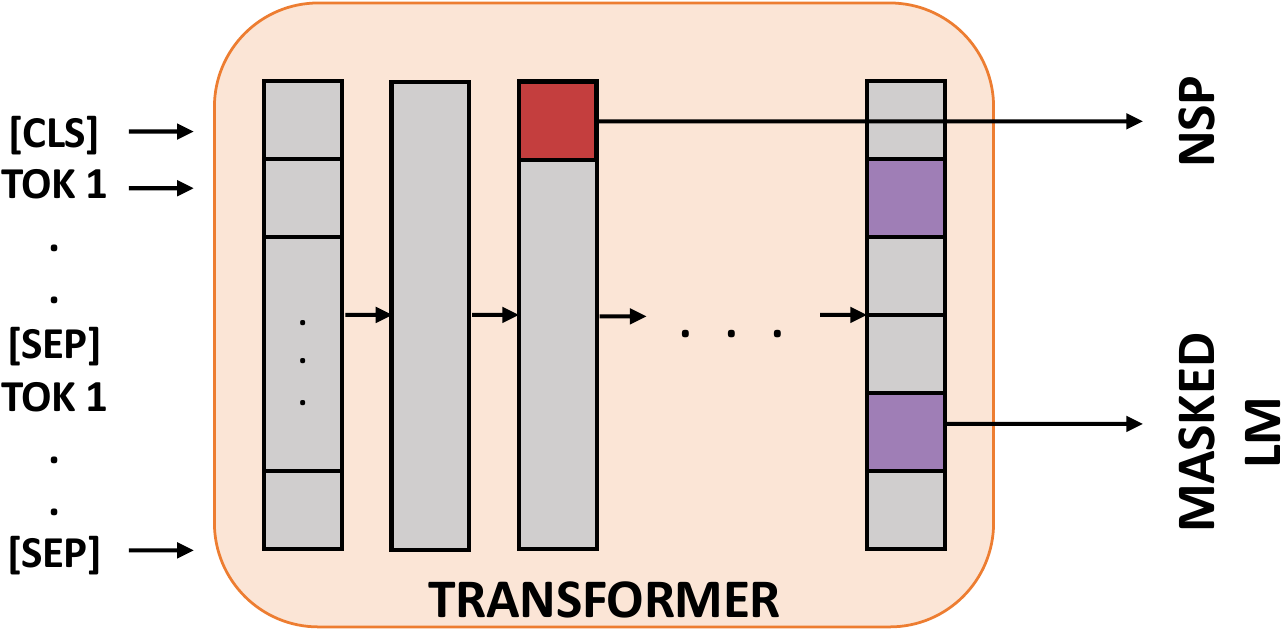}
     \caption{Lower NSP}
     \label{fig:pre-train-lowernsp}
 \end{subfigure}
 \hfill
  \begin{subfigure}[b]{0.48\textwidth}
     \centering
     \includegraphics[width=\textwidth]{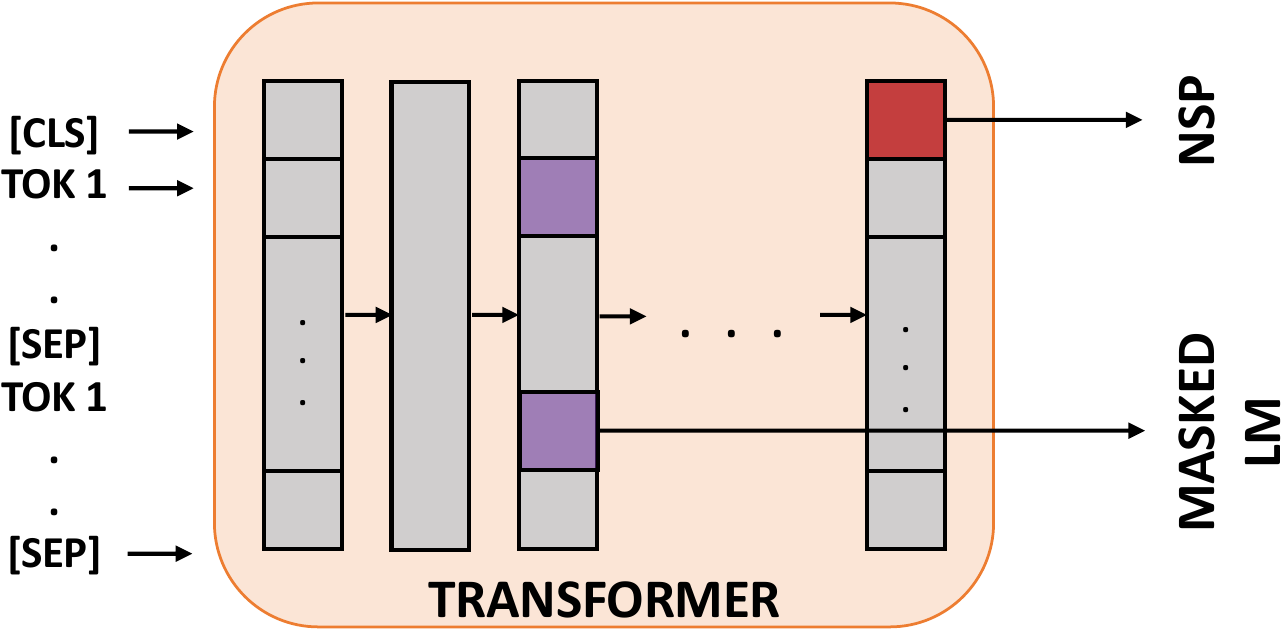}
     \caption{Lower Mask}
     \label{fig:pre-train-lowermask}
 \end{subfigure}
\caption{Different hierarchical BERT architectures. (a) using lower level embedding of \texttt{[CLS]} token for NSP classifier; (b) using lower level embeddings of masked tokens for masked LM classifier.}
\label{fig:pre-train-lower}
\end{figure}

\begin{figure}
\centering
 \begin{subfigure}[b]{0.48\textwidth}
     \centering
     \includegraphics[width=\textwidth]{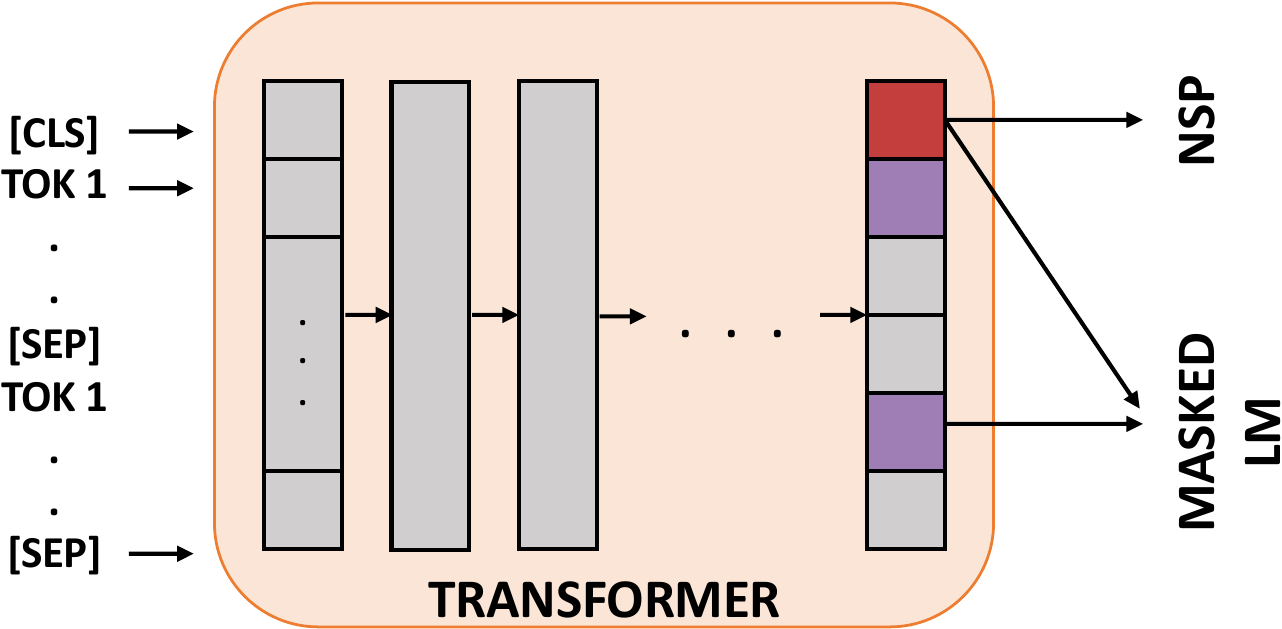}
     \caption{CLS concatenation}
     \label{fig:pre-train-clsconcat}
 \end{subfigure}
 \hfill
  \begin{subfigure}[b]{0.48\textwidth}
     \centering
     \includegraphics[width=\textwidth]{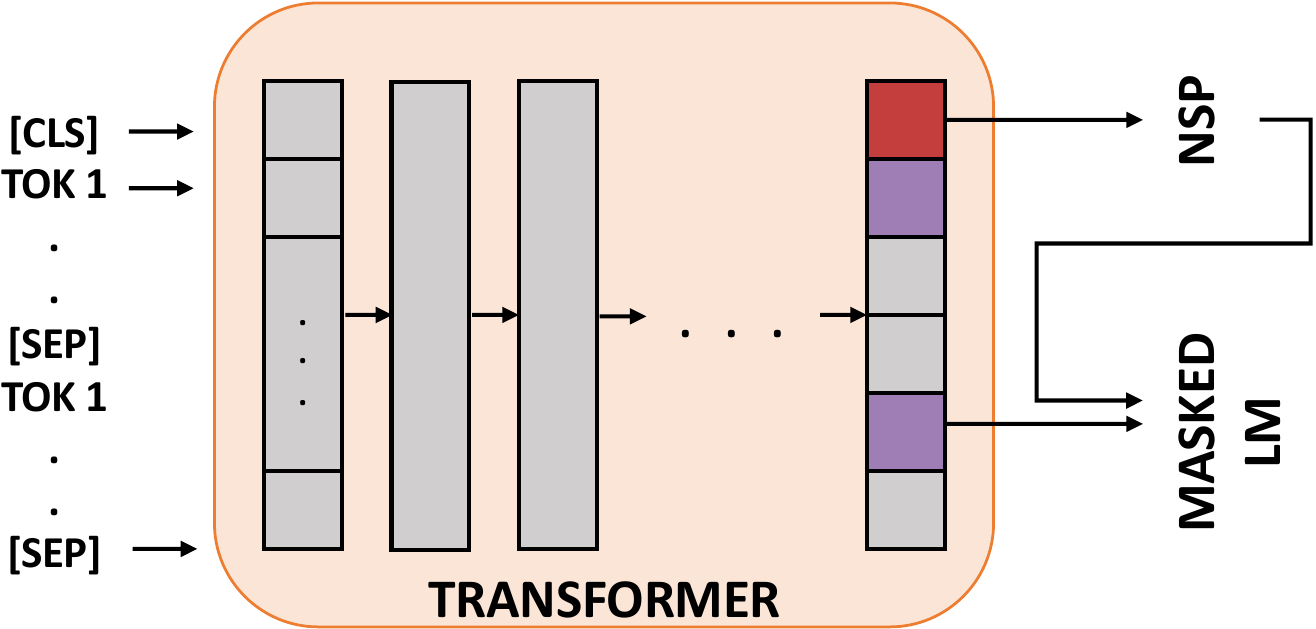}
     \caption{NSP concatenation}
     \label{fig:pre-train-nspconcat}
 \end{subfigure}
\caption{Different concatenation techniques. (a) concatenation of \texttt{[CLS]} embedding to each input of masked LM classifier; (b) concatenation of the output of NSP classifier to each input of masked LM classifier.}
\label{fig:pre-train-concat}
\end{figure}

We experimented a multitask learning approach to BERT pre-training in several ways:

\begin{enumerate}
\item NSP classifier is trained with \texttt{[CLS]} embedding from lower layers instead of the last layer. We refer to this model as Lower NSP (Figure \ref{fig:pre-train-lowernsp}). Here \texttt{[CLS]} is the start token of the input. This embedding is considered to be a sentence-level embedding \cite{devlin2018bert}. Since we do not exactly know the task hierarchy, we also do the opposite where the masked LM classifier is trained with the embeddings from lower layers, which correspond to masked tokens in the input. We refer to this model as Lower Mask (Figure \ref{fig:pre-train-lowermask}).
\item We concatenate \texttt{[CLS]} embedding, (Figure \ref{fig:pre-train-clsconcat}), or the output of the NSP classifier to the input of masked LM classifier (Figure \ref{fig:pre-train-nspconcat}). This will (1) indirectly regularize the sentence-level embedding since it will be used for both tasks (2) explicitly provide sentence-level information to masked LM classifier.
\item Motivated from probing tasks \cite{conneau2018you}, we randomly swap the order of bigrams 15\% of the time for each input and use an additional classifier, which predicts whether a token is in the right position in the sentence or not. This will force embeddings to encapsulate the word order information.
\item In our experiments for Lower NSP models, we see that the NSP classifier fits faster than the masked LM classifier. Further training will cause overfitting for the NSP classifier. Therefore we do experiments on freezing all parts that affect the NSP classifier. We refer to this model as Lower NSP-freeze.
\end{enumerate}

\begin{figure}[htbp]
\centering
 \begin{subfigure}[b]{0.2\textwidth}
     \centering
     \includegraphics[width=\textwidth]{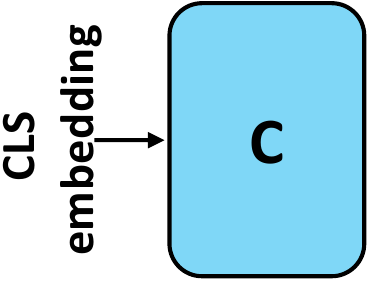}
     \caption{Sentence level tasks}
     \label{fig:fine-tune-sentence}
 \end{subfigure}
 \hfill
  \begin{subfigure}[b]{0.2\textwidth}
     \centering
     \includegraphics[width=\textwidth]{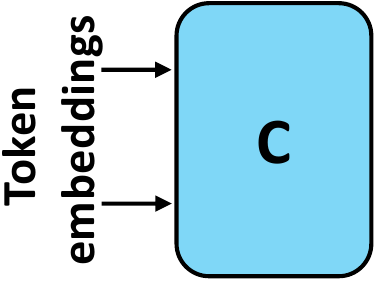}
     \caption{Token level tasks}
     \label{fig:fine-tune-token}
 \end{subfigure}
 \hfill
   \begin{subfigure}[b]{0.2\textwidth}
     \centering
     \includegraphics[width=\textwidth]{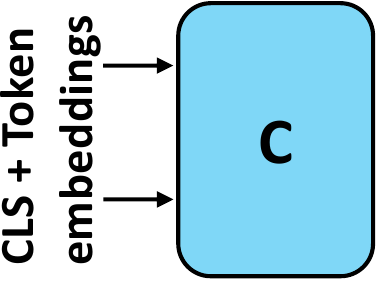}
     \caption{CLS concatenation}
     \label{fig:fine-tune-clsconcat}
 \end{subfigure}
 \hfill
    \begin{subfigure}[b]{0.2\textwidth}
     \centering
     \includegraphics[width=\textwidth]{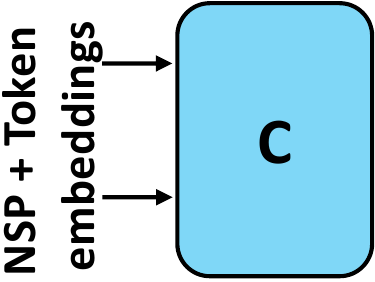}
     \caption{NSP concatenation}
     \label{fig:fine-tune-nspconcat}
 \end{subfigure}
 
\caption{Different downstream architectures. In all four figures $C$ is a classifier. (a) \texttt{[CLS]} embedding from NSP level is used as input for sentence-level tasks. (b) Embeddings from masked LM level are used as inputs for token-level tasks. In (c), \texttt{[CLS]} embedding from NSP level is concatenated to each input. In (d), NSP classifier output is concatenated to each input.}
\label{fig:fine-tune}
\end{figure}

There are several changes in the fine-tuning to accommodate the changes in the pre-training. In the original BERT structure, last layer embeddings are used for both sentence-level tasks (Figure \ref{fig:fine-tune-sentence}) and token-level tasks (Figure \ref{fig:fine-tune-token}). With our hierarchical approach, \texttt{[CLS]} embedding is selected from the layer which we train NSP classifier (Figure \ref{fig:pre-train-lowernsp}); token embeddings are selected from the layer which we train masked LM classifier (Figure \ref{fig:pre-train-lowermask}). If \texttt{[CLS]} embedding or NSP output is included in the pre-training, we also include these extra information in the fine-tuning for token-level tasks (Figure \ref{fig:fine-tune-clsconcat}, \ref{fig:fine-tune-nspconcat}).

\section{Experiments}
\label{experiments}

\subsection{Datasets}

We tested different proposed architectures on several datasets. Due to computational issues, we make pre-training on small datasets as opposed to BERT. For this, we have two different approaches. First, we used the raw text data of the downstream task for pre-training. In the second approach, we use a different raw text with similar size of the first data. We experimented on two downstream tasks: question answering (token-level task) and textual entailment (sentence-level task).

For question answering, we used SQuAD1.1 \cite{rajpurkar2016squad} and SQuAD2.0 \cite{rajpurkar2018know} datasets. SQuAD2.0 is an extended version of SQuAD1.1. It contains questions that do not have a possible answer in the given context. These two datasets are used for both pre-training and fine-tuning on downstream tasks. Additionally, we used WikiText-2 for pre-training.  We followed the same strategy as in BERT to create pre-training data from paragraphs of these three datasets separately. We pre-train the model with inputs that contain less than 128 tokens for 90\% of steps, as in BERT. However, for the rest of 10\% of steps, we used inputs containing less than 384 tokens instead of 512.

For textual entailment, we used Multi-Genre Natural Language Inference (MultiNLI) corpus \cite{williams2017broad}. This dataset does not contain paragraphs but independent sentences. Therefore we did not use this dataset for pre-training. Instead, we used the model pre-trained on the WikiText-2 dataset.

To evaluate quality of learned embeddings, we used probing tasks in \cite{conneau2018you}. We only evaluate embeddings that are pre-trained on WikiText-2 dataset since these are used for both downstream tasks. There are 10 different probing tasks which are categorized into three groups: surface information, syntactic information, and semantic information. Surface information tasks are sentence length (\textbf{SL}) and word content (\textbf{WC}). Syntactic information tasks are tree depth (\textbf{TD}), bigram shift (\textbf{BS}), and top constituents (\textbf{TC}). Semantic information tasks are tense (\textbf{T}), subject number (\textbf{SN}), object number (\textbf{ON}), semantic odd man out (\textbf{OM}) and coordination inversion (\textbf{CI}). Details of these tasks can be found in \cite{conneau2018you}.

\subsection{Pre-training and Fine-tuning Results}

We made modifications that are explained in Section \ref{methods} to the BERT base architecture. We pre-trained all variants of architectures in our pre-training datasets and fine-tuned them on downstream tasks. We select 1e-4 as the learning rate with 1e-4 weight decay for pre-training, and 1e-5 as the learning rate with no weight decay for fine-tuning. These hyperparameters are found by grid search done on BERT base architecture. We set the batch size to be 32 for inputs with length up to 128 tokens and one for longer inputs in the pre-training. Batch size for fine-tuning is set to be one for all downstream tasks. We set dropout rate to 0.1. For all experiments, Adam optimizer \cite{kingma2014adam} is used with default beta parameters and AMSGrad \cite{reddi2019convergence} option.

For Lower NSP and Lower Mask architectures, we experimented with all intermediate layers to choose the best level of selected embeddings for the lower classifier. As we used BERT base architecture, which has 12 encoder layers, there are 11 different intermediate layers. As an additional experiment, we use \texttt{[CLS]} concatenation or NSP concatenation for BERT, Lower NSP, and Lower Mask architectures. Apart from these models, we freeze NSP related parts of the model for cases in which Lower NSP classifier overfitted to data before masked LM training. To see the importance of the NSP task, we removed the NSP classifier as in RoBERTa. The best models for each architecture are fine-tuned on downstream tasks. We add three baselines:
\begin{enumerate}
\item Pre-trained BERT base model is fine-tuned.
\item BERT base architecture without any pre-training is directly fine-tuned to understand whether the performance is due to architecture or large-scale training.
\item Another modern architecture due to its success, BiLSTM, is fine-tuned with fastText \cite{bojanowski2017enriching} word embeddings.
\end{enumerate}

\begin{table}[htbp]
\centering
\begin{tabular}{l|cc|cc|cc}
\toprule
\hline
\multirow{3}{*}{\textbf{Models}} & \multicolumn{2}{c|}{\textbf{Conc.}} & \multicolumn{2}{c|}{\textbf{SQuAD}} & \multicolumn{2}{c}{\textbf{WikiText-2}} \\
 &  \multirow{2}{*}{CLS} &  \multirow{2}{*}{NSP} & SQuAD1.1 & SQuAD2.0 & SQuAD1.1 & SQuAD2.0 \\
 & & & EM / F1 & EM / F1 & EM / F1 & EM / F1 \\
 \hline
 \hline
 Pre-trained BERT & - & - & 75.9 / 85.4 & 71.1 / 73.8 & - & -  \\
\hline
\hline
 \multirow{3}{*} {BERT} & - & - & 49.5 / 60.6 & 54.8 / 57.4 & 48.1 / 60.5 & 54.7 / 57.9 \\
 & + & - & 48.1 / 59.1 & \textbf{56.7} / \textbf{59.7} & 48.8 / 60.6 & \textbf{55.0} / \textbf{58.0} \\
 & - & + & 48.7 / 60.2 & 55.5 / 58.5 & 48.2 / 59.7 & 51.2 / 53.2 \\
 \hline
\multirow{3}{*} {Lower NSP} & - & - & 47.2 / 58.3 & 56.3 / 58.8 & \textbf{51.4} / \textbf{62.7} & 54.1 / 55.9  \\
 & + & - & 47.0 / 58.5 & 55.8 / 58.3 & 46.8 / 57.6 & 53.6 / 56.6 \\
 & - & + & \textbf{52.1} / \textbf{63.4} & 54.8 / 57.3 & 46.0 / 57.6 & 54.7 / 57.8 \\
 \hline
 \multirow{3}{*} {Lower Mask} & - & - & 27.5 / 34.0 & 48.1 / 48.8 & 41.8 / 52.2 & 47.1 / 47.4  \\
 & + & - & 43.3 / 53.5 & 47.6 / 48.2 & 25.0 / 31.8 & 46.9 / 47.4 \\
 & - & + & 45.5 / 56.3 & 51.2 / 53.4 & 23.7 / 30.2 & 48.1 / 48.6  \\
 \hline
 Lower NSP-Freeze & - & - & 40.9 / 51.4 & 51.7 / 53.4 & 39.7 / 49.5 & 50.2 / 52.3 \\
\hline
Without NSP & - & - & 30.2 / 37.5 & 46.9 / 47.5 & -  & -  \\
\hline
Bigram Shift & - & - & 47.0 / 57.6 & 55.6 / 58.0 & 34.5 / 45.7 & 46.0 / 48.4  \\
\hline
BiLSTM & - & - & 22.9 / 29.3 & 46.4 / 46.7 & - & - \\
Without Pre-training & - & - & 7.4 / 12.1 & 51.0 / 51.0 & - & - \\
\hline
\bottomrule
\end{tabular}
\caption{Exact Match (EM) and F-measure (F1) results of QA classifiers on validation set. \textbf{SQuAD} and \textbf{WikiText-2} columns represent pre-training sets. \textbf{Conc.} (concatenation) column represents whether \texttt{[CLS]} embedding or NSP output is used in both pre-training and fine-tuning.} 
\label{tab:qa_results}
\end{table}

Fine-tuning results of QA classifiers on SQuAD1.1 and SQuAD2.0 validation sets are shown in Table \ref{tab:qa_results}. Here, exact match (EM) measures whether the prediction overlaps with the truth exactly. F-measure (F1) evaluates partial overlaps \cite{rajpurkar2016squad}. We see that pre-trained BERT embeddings outperform all approaches. However, we note that it is not a fair comparison since we only pre-train our models with small datasets. In the figure, BERT is the replica, except it is pre-trained on the small data. Therefore, we compare our approaches with this model. 

One obvious result is the underperformance of Lower Mask models. However, Lower NSP models show competitive performance with BERT, even better in some datasets. This might give a clue about the task hierarchy between masked LM and NSP tasks. One might expect that the masked LM task is more straightforward than the NSP task because the former makes predictions in the word-level while the latter makes predictions in the sentence-level. However, the results suggest that the masked LM task is complicated as it requires deeper layers to perform successfully. NSP task can be done without encoding the full knowledge of each word. Unlike RoBERTa, the performance drops when we remove NSP loss. The reason for this might be pre-training with small data; the model may not need sentence-level task when it has access to many word combinations. Instead of removing NSP loss, one promising direction can be training NSP task at lower layers for the large-scale training.

Bigram Shift model performs slightly worse than other models when the pre-training is done on SQuAD datasets. One possible explanation is that the question answering problem does not require the word order information. On WikiText-2, the results are dramatically worse. This might be due to optimization; a better hyperparameter search can increment the performance.

Self-supervised pre-training shows its contribution once more. Even when pre-training is done on the downstream data with a poorly constructed architecture (Lower Mask), it performs better than the original transformer without any pre-training. In Table \ref{tab:qa_results}, without pre-training model has 51.0 F1 score on SQuAD2.0. However, this is not a good model as it predicts that all question-paragraph pairs are impossible to answer (51\% of examples in the validation set are impossible to answer). Furthermore, all models outperform BiLSTM with fastText. Notice that, fastText is also a model pre-trained on large-scale data. A transformer pre-trained directly on downstream data has better representations for the specific downstream tasks than BiLSTM with fastText.

\begin{table}[htbp]
\centering
\begin{tabular}{l|cc|cc}
\toprule
\hline
\multirow{2}{*}{\textbf{Models}} & \multicolumn{2}{c|}{\textbf{Pre-training}} & \multicolumn{2}{c}{\textbf{Accuracy}} \\
 & CLS & NSP & Matched & Mismatched \\
\hline
\hline
Pre-trained BERT & - & - & 84.4 &  84.9 \\
\hline
\hline
\multirow{3}{*} {BERT} & - & - & \textbf{71.7} & \textbf{72.5} \\
& + & - & 70.2 & 70.8 \\
& - & + & 71.1 & 71.3 \\
\hline
\multirow{3}{*}{Lower Mask} & - & - & 69.9 & 70.7 \\
& + & - & 68.5 & 69.2 \\
& - & + & 67.1 & 67.9 \\
\hline
\multirow{3}{*}{Lower NSP} & - & - & 69.1 & 70.2 \\
& + & - & 70.3 & 71.9 \\
& - & + & 70.5 & 71.3 \\
\hline
Lower NSP-Freeze & - & - & 45.2 & 45.3 \\
\hline
Bigram Shift & - & - & 69.8 & 70.7 \\
\hline
BiLSTM & - & - & 66.8 & 67.3 \\
Without Pre-training & - & - & 61.2 & 61.4 \\
\hline
\bottomrule
\end{tabular}
\caption{Accuracy results on MultiNLI validation sets. \textbf{Pre-training} column represents whether \texttt{[CLS]} embedding or NSP output is used in pre-training.}
\label{tab:mnli}
\end{table}

Results on the MultiNLI corpus for both matched and mismatched validation sets are shown in Table \ref{tab:mnli}. Matched set contains examples from the same source as the training set; mismatched set contains examples from different sources. We cannot use concatenated parts for fine-tuning, unlike question answering. Only \texttt{[CLS]} embedding from NSP level is used for fine-tuning.

As in QA results, pre-training on small data has a bootstrapping effect compared with the model without pre-training. All models except Lower NSP-Freeze outperforms BiLSTM. This shows that the transformer aggregates sentence-level information better than BiLSTM. This is an interesting result since BiLSTM sees each word explicitly to encode sentence-level embedding while the transformer encodes this implicitly via the NSP task. While BERT has the best performance, Lower NSP is also competitive. Unlike question answering, Lower Mask and Bigram Shift models have better performance.

\subsection{Ablation Studies}

\begin{table}[htbp]
\centering
\begin{tabular}{l|cc|cc|cc|cc}
\toprule
\hline
\multirow{3}{*}{\textbf{Models}} & \multicolumn{2}{c|}{\textbf{PT}} &  \multicolumn{2}{c|}{\textbf{FT}} & \multicolumn{2}{c|}{\textbf{SQuAD}} & \multicolumn{2}{c}{\textbf{WikiText-2}} \\
 & CLS & NSP & CLS & NSP & SQuAD1.1 & SQuAD2.0 & SQuAD1.1 & SQuAD2.0 \\
 & & & & & EM / F1 & EM / F1 & EM / F1 & EM / F1 \\
 \hline
 \hline
\multirow{5}{*} {BERT} & - & - & - & - & 49.5 / 60.6 & 54.8 / 57.4 & 48.1 / 60.5 & 54.7 / 57.9  \\
\cline{2-9}
 & - & - & + & - & 50.0 / 61.6 & 55.1 / 58.4 & 48.0 / 60.7 & 54.9 / 58.2 \\
 & - & - & - & + & 50.4 / 61.7 & 56.5 / 59.5 & 47.4 / 60.1 & \textbf{55.6} / \textbf{58.9} \\
 \cline{2-9}
 & + & - & + & - & 48.1 / 59.1 & \textbf{56.7} / \textbf{59.7} & 48.8 / 60.6 & 55.0 / 58.0 \\
 & - & + & - & + & 48.7 / 60.2  & 55.5 / 58.5 & 48.2 / 59.7 & 51.2 / 53.2 \\
\hline
\hline
\multirow{7}{*} {Lower NSP} & - & - & - & - & 47.2 / 58.3 & 56.3 / 58.8 & \textbf{51.4} / \textbf{62.7} & 54.1 / 55.9  \\
\cline{2-9}
 & - & - & + & - & 47.9 / 59.5 & 56.7 / 59.6 & 50.7 / 62.1 &  54.0 / 57.1\\
 & - & - & - & + & 48.3 / 59.4 & 56.3 / 59.3 & 50.6 / 62.3 & 54.6 / 57.7 \\
  \cline{2-9}
 & + & - & + & - & 47.0 / 58.5 & 55.8 / 58.3 & 46.8 / 57.6 & 53.6 / 56.6 \\
 & - & + & - & + & \textbf{52.1} / \textbf{63.4} & 54.8 / 57.3 & 46.0 / 57.6 & 54.7 / 57.8 \\
  \cline{2-9}
 & + & - & - & - & 47.1 / 58.3 & 55.6 / 58.3 & 46.9 / 58.0 & 53.4 / 56.3 \\
 & - & + & - & - & 51.0 / 62.8 & 54.2 / 56.9 & 46.5 / 58.1 & 55.7 / 58.4 \\
 \hline
 \bottomrule
\end{tabular}
\caption{Exact Match (EM) and F-measure (F1) results about using concatenated parts in pre-training (\textbf{PT}) and/or fine-tuning (\textbf{FT}).}
\label{tab:qa_ablation}
\end{table}

To evaluate the effect of concatenated parts (\texttt{[CLS]} embedding or NSP output), we also test the following architectures: (1) use in pre-training (2) use in fine-tuning (3) use in both pre-training and fine-tuning (4) not using. QA classifier results for these experiments are shown in Table \ref{tab:qa_ablation}. Even if the model is pre-trained without using concatenated parts, using these extra inputs in fine-tuning slightly increases the performance. In some cases, the peak performance is achieved by using the concatenated parts in both pre-training and fine-tuning.

\subsection{Probing Tasks}

\begin{table}[htbp]
\centering
\begin{tabular}{l|cc|cccccccccc}
\toprule
\hline
\multirow{2}{*}{\textbf{Models}} & \multicolumn{2}{c|}{\textbf{PT}} & \multirow{2}{*}{\textbf{SL}} & \multirow{2}{*}{\textbf{WC}} & \multirow{2}{*}{ \textbf{TD}} & \multirow{2}{*}{ \textbf{BS}} &  \multirow{2}{*}{\textbf{TC}} & \multirow{2}{*}{\textbf{T}} & \multirow{2}{*}{\textbf{SN}} & \multirow{2}{*}{\textbf{ON}} & \multirow{2}{*}{\textbf{OM}} & \multirow{2}{*}{\textbf{CI}} \\
 & CLS & NSP  \\
 \hline
 \hline
PT. BERT & - & - &  68.3 & 32.4 & 34.3 & 86.5 & 75.2 & 88.7 & 83.0 &	77.8 & 64.3 & 74.4 \\
\hline
\hline
\multirow{3}{*}{BERT} & - & - & 83.8 &	9.6 &	 36.0 &	62.5 & 70.1 & \textbf{75.9} & 74.7 & 69.9 & 49.7 & \textbf{59.8} \\
& + & - & 75.7 & 10.1 & 34.6 & 57.9 & 63.6 & 70.2 & 73.8 & 68.9 & 50.5 & 55.6 \\
& - & + & 84.9 & \textbf{12.4} & 34.9 & 60.4 & 60.7 & 72.2 & 73.2 &	69.2 & 49.9 & 58.6 \\
\hline
\multirow{3}{*}{Lower Mask} & - & - & 73.2 & 2.2 &	29.0 & 56.1 &	 61.3 & 73.6 & 73.0 & 67.1 & 50.2 &	56.8 \\
& + & - & 43.8 & 1.8 & 23.5 & 54.1 & 30.7 & 70.5 & 62.2 & 58.2 & 49.8 & 51.5 \\
& - & + & 41.7 & 0.8 & 21.3 & 52.1 & 26.7 & 66.6 &	 62.3 & 58.0 & 49.9 & 51.7 \\
\hline
\multirow{3}{*}{Lower NSP} & - & - & 91.0 & 5.6 & 31.7 & 52.9 &	57.5 & 71.7 & 71.9 & 66.4 & \textbf{51.1} & 56.0 \\
& + & - & 86.1 & 11.1 & 34.5 & 60.0 & 67.9 & 74.9 & \textbf{75.4} &	\textbf{71.5} & 50.8 &	 57.2  \\
& - & + & 90.1 & 1.1 & 31.0 & 57.6 & 71.8 & 70.9 & 72.7 & 65.5 & 50.3 & 57.6 \\
\hline
L. NSP-Freeze & - & - & \textbf{93.2} & 1.5 & 33.7 & 53.8 & \textbf{72.3} & \textbf{75.9} & 75.2 & 68.8 & 50.3 & 57.2 \\
\hline
Bigram Shift & - & - & 88.6 & 3.8 & \textbf{37.2} &	\textbf{70.3} & 65.6 & 70.5 & 72.7 & 68.2 & 49.9 & \textbf{59.8} \\
\hline
\bottomrule
\end{tabular}
\caption{Accuracy results of probing tasks. \textbf{PT} (pre-training) column represents whether \texttt{[CLS]} embedding or NSP output is used in pre-training.}
\label{tab:prob}
\end{table}

For probing tasks, a multi-layer perceptron (MLP) with two hidden layers with 128 units is used. Here, the transformer part is frozen, and we only train the probing task classifier. The learning rate is set to be 1e-3 with no weight decay. The batch size is 32. The classifier uses \texttt{[CLS]} embedding from the NSP level as input.

The results of probing tasks are shown in Table \ref{tab:prob}. These results suggest the followings:

\begin{itemize}
\item In sentence length and tree depth tasks, pre-trained BERT performs worse than some models. This is quite surprising considering its large-scale training.
\item Lower NSP models perform better than Lower Mask models and better than BERT for some tasks. We can say that Lower NSP constructs more informative sentence-level representation than the others.
\item Bigram shift model achieves good results for tasks that require order information (tree depth, bigram shift, coordination inversion). While it is not as good as other models in downstream tasks, including order information might be beneficial for other tasks.
\end{itemize}

\section{Conclusion}
\label{conclusion}

We proposed to pre-train BERT with a hierarchical multitask learning approach. Our results on restricted data (due to computational resources) show that this approach achieves better or equal performance. We incorporate sentence-level information to solve word-level tasks. This also shows a slight increment in performance. We propose an additional pre-training task, bigram shift, which causes embeddings to contain word order information. 

We believe that implementing these techniques to large-scale training will further advance the state-of-the-art. Probing tasks show that different training techniques lead embeddings to contain different linguistic properties. This is an essential point since there are various problems in the NLP domain that require different needs. Therefore selecting an appropriate pre-training strategy is an important factor.

\section*{Acknowledgments}
The numerical calculations reported in this work were partially performed at TUBITAK ULAKBIM, High Performance and Grid Computing Center (TRUBA resources) and TETAM servers.

\bibliographystyle{coling}
\bibliography{ref}

\end{document}